\documentclass[conference]{IEEEtran}
\IEEEoverridecommandlockouts

\usepackage{cite}
\usepackage{amsmath,amssymb,amsfonts}
\usepackage{algorithmic}
\usepackage{graphicx}
\usepackage{textcomp}
\usepackage{xcolor}
\usepackage{nccfoots}
\usepackage{booktabs} 

\usepackage[colorlinks = true,
            linkcolor = blue,
            urlcolor  = blue,
            citecolor = red]{hyperref}

\usepackage{amsmath,amsfonts,bm}

\def\eqref#1{equation~\ref{#1}}

\def\1{\bm{1}}

\DeclareMathAlphabet{\mathsfit}{\encodingdefault}{\sfdefault}{m}{sl}
\SetMathAlphabet{\mathsfit}{bold}{\encodingdefault}{\sfdefault}{bx}{n}

\DeclareMathOperator*{\argmax}{arg\,max}

\newcommand{\chapternote}[1]{{%
  \let\thempfn\relax
  \footnotetext[0]{\emph{#1}}
}}

\begin{document}

\title{Flexible Multi-task Networks\\ by Learning Parameter Allocation}

\author{\IEEEauthorblockN{Krzysztof Maziarz\textsuperscript{*}}
\IEEEauthorblockA{\textit{Jagiellonian University}}
\and
\IEEEauthorblockN{Efi Kokiopoulou}
\IEEEauthorblockA{\textit{Google}}
\and
\IEEEauthorblockN{Andrea Gesmundo}
\IEEEauthorblockA{\textit{Google}}
\and
\IEEEauthorblockN{Luciano Sbaiz}
\IEEEauthorblockA{\textit{Google}}
\and
\IEEEauthorblockN{Gabor Bartok}
\IEEEauthorblockA{\textit{Google}}
\and
\IEEEauthorblockN{Jesse Berent}
\IEEEauthorblockA{\textit{Google}}
}


\maketitle

\Footnotetext{*}{Work done during an internship at Google. The author is now at Microsoft Research. Correspondence to \texttt{krzysztof.maziarz@microsoft.com}}

\begin{abstract}
This paper proposes a novel learning method for multi-task applications.
Multi-task neural networks can learn to transfer knowledge across different tasks by using parameter sharing. However, sharing parameters between unrelated tasks can hurt performance.
To address this issue, we propose a framework to learn fine-grained patterns of parameter sharing.
Assuming that the network is composed of several components across layers, our framework uses learned binary variables to allocate components to tasks in order to encourage more parameter sharing between related tasks, and discourage parameter sharing otherwise.
The binary allocation variables are learned jointly with the model parameters by standard back-propagation thanks to the Gumbel-Softmax reparametrization method.
When applied to the Omniglot benchmark, the proposed method achieves a $17\%$ relative reduction of the error rate compared to  state-of-the-art.
\end{abstract}


\section{Introduction}\label{sec:introduction}

Multi-task learning \cite{Caruana-1998, Caruana-1993} based on neural networks has attracted a lot of research interest in the past years, and has been successfully applied to several application domains, such as recommender systems \cite{BaBeCa-2016}, real-time object detection \cite{Fast-R-CNN} and learning text representations \cite{DBLP:journals/corr/abs-1907-04829,DBLP:journals/corr/abs-1907-12412}.
For instance, a movie recommendation system may optimize not only the likelihood of the user clicking on a suggested movie, but also the likelihood that the user is going to watch it.

The most common architecture used in practice for multi-task learning is the so-called \textit{shared bottom}, where the tasks share parameters in the early layers of the model, which are followed by task-specific heads.
However, as our experiments on synthetic data show, when the tasks are unrelated, parameter sharing may actually hurt individual tasks performance. Therefore, resorting to flexible parameter sharing becomes very important. This can be achieved by manually trying several different static sharing patterns. However, this option is not scalable, since it requires significant effort. Instead, an approach where the sharing pattern is learned and adapted to the task relatedness is preferable. 

In this work, we introduce a novel method that learns the sharing pattern jointly with the model parameters using standard back-propagation.
Assuming that the network consists of several layers, where each layer consists of several components, the core idea of the proposed method is to learn, for each component, a set of binary allocation variables indicating which tasks use this component.
We rely on the Gumbel-Softmax reparameterization method \cite{JangGuPoole.2017} in order to train these binary variables jointly with the parameters of the components.

We provide experiments on both synthetic tasks and commonly used benchmarks, showing that the proposed framework can adapt the sharing pattern to the task relatedness, outperforming strong baselines and previous state-of-the-art methods. In summary, the paper contributions are the following:
\begin{itemize}
    \item We analyze positive and negative transfer effects on synthetic tasks. This analysis motivates the need for task-dependent parameter sharing in multi-task networks.
    \item We propose a novel multi-task learning method based on Gumbel binary variables. It allows for learning flexible parameter sharing which adapts to the task relatedness, and can be optimized with standard back-propagation.
    \item We show that our method can be used to produce sparse task embeddings by concatenating the binary allocation patterns. We provide experimental evidence that such embeddings reflect task relatedness.
\end{itemize}
The source code implementing the proposed method and the benchmarks described in this paper is publicly available under \href{https://drive.google.com/open?id=1vrXONNr_SBzQw81gZlqsWMslkFmzy2HK}{this link}.

\begin{figure*}[ht]
\begin{center}
\includegraphics[width=0.9\linewidth]{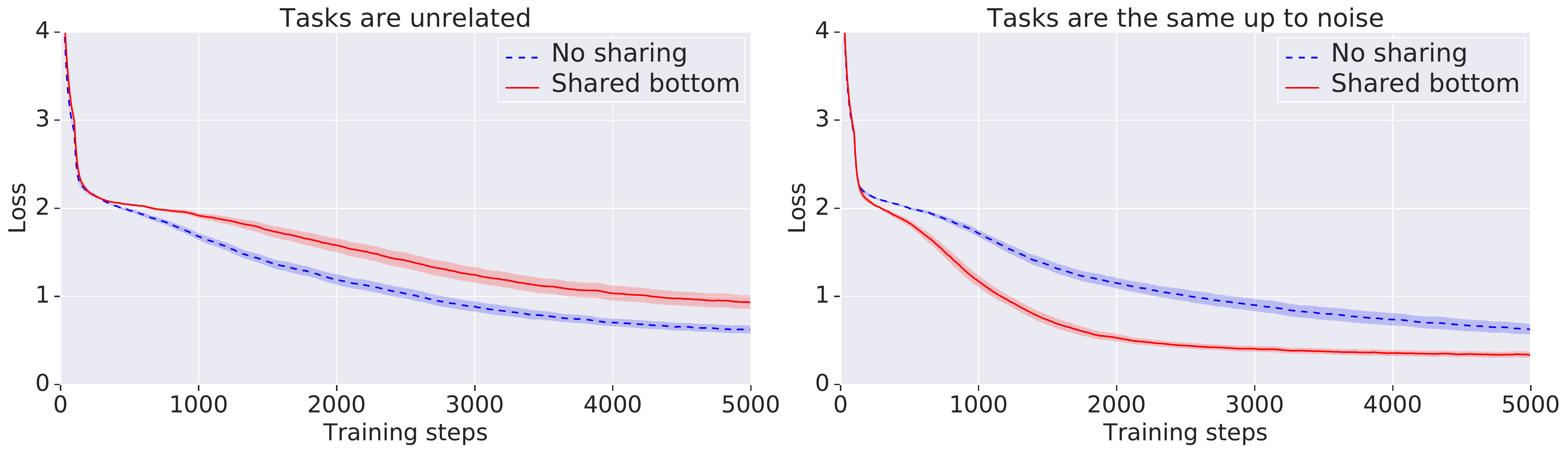}
\end{center}
\caption{Comparison of `shared bottom' and `no sharing' patterns for unrelated tasks (left) and almost equal tasks (right). The plots show loss over time (averaged over tasks and smoothed over a window of $100$ steps). We ran each experiment $30$ times, and the shaded area shows the $90\%$ confidence interval.}\label{fig:pos_neg_transfer}
\end{figure*}

The rest of the paper is organized as follows.
In Section \ref{sec:positive_negative_transfer} we provide experiments on synthetic data, showing that negative transfer may occur in practice, which motivates the need for flexible parameter sharing. In Section \ref{sec:proposed_framework} we introduce in detail the proposed framework, and discuss related works in Section~\ref{sec:related_work}. Section \ref{sec:experiments} shows the experimental results, and Section \ref{sec:conclusions} summarizes our paper and outlines future work.

\section{Positive and negative transfer}\label{sec:positive_negative_transfer}

We start with a practical example, showing that besides positive transfer, negative transfer may occur as well. That is, when the tasks are unrelated, allowing them to interact in a bigger model instead of training them separately harms the model performance. 
To show that both positive and negative transfer occurs, we generate two synthetic tasks, where the task relatedness $\rho$ can be explicitly controlled. Our synthetic data generation process is based on that of~\cite{MMoE-KDD-2018}, and we describe it in detail in Appendix
~\ref{appendix:synthetic_data_generation}.
We consider two edge cases: two unrelated tasks ($\rho = 0$), and two tasks that are the same up to noise ($\rho = 1$).

We create a simple multi-task network. This network architecture consists of four parallel components, and each component contains a stack of fully connected layers. Each input example can be passed through any subset of the four parallel components; the outputs of the components are averaged before being passed to a task-specific linear head. We chose this network to have low enough capacity, so that there is visible competition between tasks. For more information about the architecture, please refer to Appendix~\ref{appendix:positive_negative_transfer_arch}.

For this experiment, we use two hard-coded sharing patterns. The \textit{`shared bottom'} pattern means that both tasks use all four components, while \textit{`no sharing'} means that each task is assigned two components out of four. Therefore, in the `no sharing' pattern, the tasks are completely independent.
Note that regardless of the pattern, the total amount of parameters in the  model remains the same; the only difference is in which parameters get used by which tasks. In other words, `no sharing` corresponds to imposing a constraint that the network should be evenly divided between the tasks, while `shared bottom' leaves that to the optimization algorithm to decide.

We run four experiments: one for every combination of sharing pattern (`shared bottom' and `no sharing'), and task relatedness ($\rho \in \{0, 1\}$). For each experiment, we report the L2 loss over time, averaged over the two tasks. The results are shown in Figure~\ref{fig:pos_neg_transfer}. Since for `no sharing' there is no interaction between the tasks, the average loss behaves in the same way irrespective of the task relatedness. For `shared bottom', both tasks are allowed to update all parameters, and we see that while it improves performance if the tasks are related, it actually hurts for two completely unrelated tasks.

Motivated by this example, we argue that general multi-task models should be able to learn flexible sharing patterns that can adapt to the task relatedness. In Section~\ref{sec:proposed_framework}, we introduce a framework which can learn such flexible sharing patterns, including but not limited to `no sharing' and `shared bottom'.

\section{Framework}\label{sec:proposed_framework}

\begin{figure*}[!t]
\begin{center}
\includegraphics[width=0.9\linewidth]{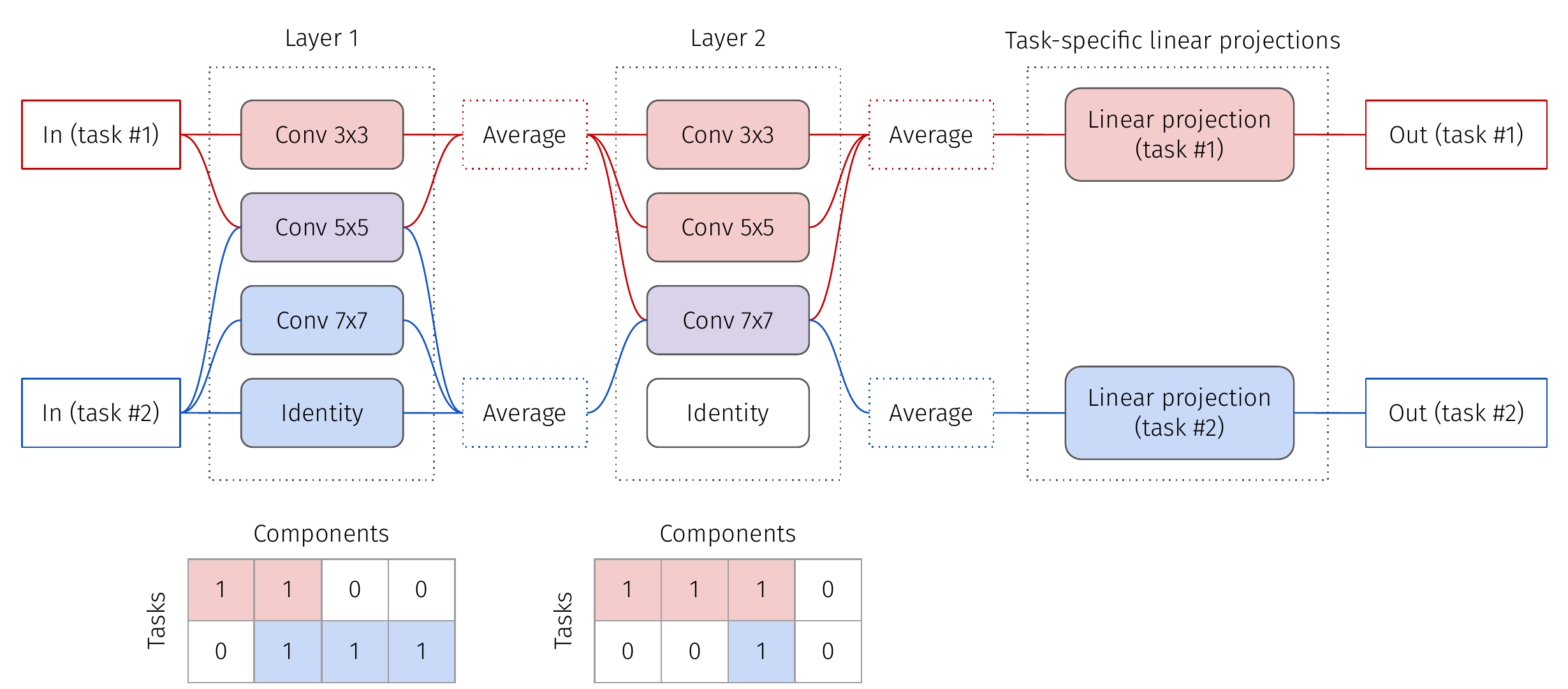}
\end{center}
\caption{An example network with two tasks. Some components are used by both tasks (purple), some by only one of the tasks (red or blue, respectively), and one identity component is completely unused (white). Below each layer we show the corresponding allocation matrix.}\label{fig:routing_model_specific}
\end{figure*}

We design our method so that it is able to learn to flexibly allocate components to tasks in a large modular network. In Figure~\ref{fig:routing_model_specific} we show an example, containing two tasks and two modular layers, followed by task-specific heads. Each task selects a subset of components within every layer. For every layer, we encode the component allocation as a binary matrix. Concretely, if there are $T$ tasks, and $C_l$ components in the $l$-th layer, the matrix for that layer has shape $T$ x $C_l$, where the $(i,~j)$ entry is $1$ if the $j$-th component is allocated to the $i$-th task. We show these binary allocation matrices at the bottom of Figure~\ref{fig:routing_model_specific}.

Our goal is to learn the binary allocation matrices in the way that maximizes the average per-task performance of the model.
Note that the set of components used by a given input is conditioned on the task id only, which implies that all samples from the same task will go through the same part of the network. We refer to our framework as the \textit{Gumbel-Matrix} framework.
At training time our method samples many different allocation matrices, but after training a single binary matrix is selected for every layer.

\subsection{Training}

For each layer, we want to maintain a probability distribution over all possible binary allocations. To make the problem tractable, we assume this distribution to be factorized, and explicitly maintain a matrix of probabilities for each component-task pair. Each probability is represented as a pair of two complementary logits. To perform a forward pass, we sample all binary connections independently according to the probability matrix. In principle, it is possible that for a given task all allocation variables for some layer are sampled to $0$, meaning that no component in that layer is assigned to the task. In that case, we assume the output of the layer to be a zero vector, independently of the input. In practice, we found that this happens very rarely, and mostly at the beginning of training, since usually one of the allocation probabilities quickly becomes close to $1.0$. Therefore, we did not try to devise a method which would artificially prevent an all-zeros allocation pattern from being sampled.

To initialize our method, we have to set the allocation probabilities to some initial values. In principle, it is possible to introduce prior knowledge, and set these probabilities in a way that encourages or discourages certain patterns. However, here we consider the most general approach, where all probabilities are initialized to the same constant value $p_{\rm init}$. Setting $p_{\rm init} = 0.5$ gives the highest entropy, and corresponds to the weakest prior, therefore we consider it as a default choice. However, in our experiments with large and deep networks, it is often beneficial to set $p_{\rm init}$ closer to $1.0$, in order to enhance the trainability of the components and to stabilize the initial learning phase.

In the backwards pass, only the components that are activated will get gradients, as the inactive components do not contribute to the final output of the network. However, in order to get a gradient for the allocation probabilities, we would have to back-propagate through sampling. In the next section, we describe a method which we use to accomplish that.

\subsection{Propagating gradients through binary samples}
In order to get gradients to the allocation probabilities, we follow \cite{JangGuPoole.2017}, and reparameterize sampling from a Bernoulli distribution using the Gumbel-Softmax trick.
The Gumbel distribution can be defined by the following forward sampling procedure:
\begin{equation*} 
u \sim \text{Uniform}(0, 1) \Rightarrow g = -\log(-\log(u)) \sim \text{Gumbel}.
\end{equation*}
Instead of using the logits to directly sample a binary value, we add independent noise from the Gumbel distribution to each of the logits, and then select the binary value with the highest logit (i.e. argmax) as the sample $z$. Formally, to sample from Bernoulli($p$), we use the following procedure. Let $\pi = [p, 1 - p]$; we draw $g_0$ and $g_1$ independently from the Gumbel distribution, and produce the sample $z$ as
\begin{equation*}
z = \argmax_{i \in \{0,1\}} v_i, \text{where}   ~v:=  \log(\pi) + [g_0, g_1].
\end{equation*}
The argmax operation is not differentiable, but it can be approximated by a softmax with annealing temperature. Therefore, on the forward pass, we use the argmax to obtain a binary connection value, while on the backwards pass, we approximate it with softmax, similarly to \cite{SpotTune-CVPR-2018}. This approach is known as the \textit{Straight-Through Gumbel-Softmax estimator} \cite{JangGuPoole.2017}.
Note that the backwards pass actually requires all components to be evaluated, irrespective of whether they are used in the forward pass or not. Therefore, if an allocation is sampled to be inactive, then the corresponding component will not get gradients, but its output will be used to compute the gradient for the allocation probability.

\subsection{Inference}

At inference time, it is possible to follow the same procedure as at training time, i.e. sample the connection pattern for every test batch. In our experiments, we found that this works well and does not introduce a large amount of noise in the evaluation result, since the allocation probabilities naturally tend to converge to either $0.0$ or $1.0$ during training. An alternative approach is to fix the allocations to their maximum likelihood variants, and use that pattern for every forward pass. We do that in the evaluation phase of all of our experiments, since we believe this is closer to how the Gumbel-Matrix framework should be used in practice. Note that for the maximum likelihood approach, we can discard all allocation probabilities after training has completed. The probabilities are used only to describe how to select a part of the network for each task.

\subsection{Budget penalty}
We have found that Gumbel-Matrix method generally trains well in its vanilla form. However, we note some ways in which it is possible to alter its default behavior. For example, one might want to learn an allocation pattern with a certain degree of sparsity. To that end, we introduce the \textit{budget penalty}, which penalizes the model from exceeding a given computational budget. We define the budget as a maximum percentage of active allocations. Since we explicitly maintain all allocation probabilities, by averaging them out over all layers we obtain the \textit{expected} fraction of active allocations $e_c$. Therefore, we can set a budget $b \in (0, 1\rangle$, corresponding to the maximum allowed fraction of active allocations, and define the \textit{budget auxiliary loss} as: $\lambda \max(0, e_c - b)$, where $\lambda$ is a constant that controls the strength of the penalty. For a sufficiently large $\lambda$, this penalty can be viewed as a hard constraint in practice.

\subsection{Sparse task embeddings}\label{subsec:embeddings}

Finally, we note that our framework can be used to produce sparse embeddings for the tasks, and uncover latent clusters of related tasks.

Let us denote the total number of components across all layers of a model by $C = \sum_l C_l$. After training with the Gumbel-Matrix framework,
for each task, we collect together all the binary allocation vectors from all layers, describing which components are allocated to this task. This results into a (concatenated) binary vector of total dimension $C$ for each of the $T$ tasks. This process can be understood as embedding all tasks into a common shared space; in this space, we can compare tasks by computing the distance between their embedding vectors. Note that for a fixed pair of tasks, higher similarity of their corresponding embedding vectors implies higher level of parameter sharing.

Based on the observation above, we conjecture that these embeddings can be used to discover latent clusters within a set of tasks. We verify this conjecture empirically in Section~\ref{subsec:clusters}.

\section{Related work}\label{sec:related_work}

Traditionally, work on multi-task learning includes hand-designing the sharing pattern, in order to strike a good balance between shared and task-specific parameters. 
Going beyond static patterns, our method is more related to recent works in multi-task learning, where the parameter sharing pattern is dynamically learned together with the model parameters. These works can be mainly divided based on the underlying algorithm  used to learn the pattern.

Some methods cast learning the pattern as a Reinforcement Learning problem. The authors in \cite{RoutingNetworks18, RoutingNetsChallenges} propose a framework based on multi-agent Reinforcement Learning, where the positive-negative transfer problem is taken care of by finding a Nash equilibrium. In contrast, our work does not rely on Reinforcement Learning, and can be trained with standard back-propagation.

Many other works use the Sparsely-Gated Mixture-of-Experts \cite{Shazeer-MoE-2017}, initially developed for a single-task model. This idea is extended in \cite{MMoE-KDD-2018} by introducing a separate gating function per task, and in \cite{Diversity&Depth-ICLR-2019} by using architecturally diverse experts and increasing the routing depth. However, \cite{Diversity&Depth-ICLR-2019} report that their models are often hard to train. In contrast, we show that our method can improve accuracy even when routing in large and deep neural networks.

In order to learn the sharing pattern, one can also use evolutionary algorithms. \cite{Fernando2017-Pathnet} use evolution to find task-specific paths through a shared computational graph.

Some methods use soft combinations instead of discrete routing decisions. Cross-stitch networks \cite{CrossStichNetworks} use single-task models that are stitched together with cross-stitch units, which learn shared representations by linearly combining intermediate representations from the single task models. \cite{MeyerMiikk-ICLR-2018} learn a soft ordering of shared layers, where different tasks use the same blocks, but composed in a different soft order.

Sub-Network Routing \cite{SNR-AAAI19} proposes a routing method for multi-task networks, but it does not learn task-dependent routing, as the same pattern is applied to all the tasks.

The authors in \cite{ManyTaskLearning} propose a task routing layer for many task learning, which is a special case of multi-task learning where more than 20 tasks are performed by a single model. The task routing layer relies on binary random masks, which are created when the model is instantiated, but the masks are not trainable and they remain fixed throughout the training process. The amount of parameter sharing across tasks is governed by a single scalar knob.

Additionally, some works use concepts similar to ones used in this paper, but not for multi-task learning. In particular, \cite{SpotTune-CVPR-2018} also rely on the Gumbel-Softmax trick to learn binary variables, with a focus on fine-tuning for transfer learning applications. The binary variables are used to decide which layers of a pre-trained model should be fine-tuned on the target task.
Another related method \cite{BengBaPiPre.2016} learns binary variables that mask the outputs of each layer, conditioning on the activations of the previous layers. In order to learn these binary variables, the REINFORCE algorithm \cite{REINFORCE-1992} is used.
A similar notion of adaptive inference graphs has been proposed in \cite{ConvNet-AIG-19} that uses convolutional neural nets for image classification based on a ResNet-type of architecture, where some layers are skipped using learned gating functions.
However, note that these methods have not been designed for multi-task learning.

Finally, our approach is related to methods for Neural Architecture Search (NAS) \cite{ZophLe16,ReMoSeSaSuTaLeKu17,maziarz2018evolutionary,white2019bananas}, which automatically design neural network architectures for a given task.
Our method is of similar spirit as the efficient NAS methods \cite{ENAS,DARTS} in the sense that the architecture parameters are jointly optimized with the model parameters.
Our method searches for a multi-task architecture that learns a flexible parameter sharing pattern according to the task relatedness, and uses a simpler architecture encoding based on binary variables.

\section{Experiments}\label{sec:experiments}

In all our experiments, we impose an additional constraint that each input batch contains samples for only one task. Since the routing is conditioned only on the task, this allows us to sample the allocation pattern once per forward pass. To train the network in a multi-task setting, we draw one batch of input samples per task, pass them through the network in random order, and repeat that process for a predefined number of steps.

To start, we have tested our method on the same synthetic data as in Section~\ref{sec:positive_negative_transfer}. We provide the experimental results in Appendix~\ref{appendix:synthetic_data_gumbel_matrix}.

\subsection{MNIST}

\paragraph{Experimental setup}
To test our method in a controlled environment where we know which pairs of tasks are more related, we create the following \textit{4-MNISTs setup} based on the MNIST dataset.

We first define the \textit{MNIST-rot} task, by taking the input-output pairs of MNIST, and rotating all input images clockwise by $90$ degrees.
We run experiments on four tasks, where the first two tasks are copies of MNIST, and the next two are copies of MNIST-rot. Note that two copies of the same task have the same training and test datasets, but the order of training examples is different.
In order to make the setup difficult, we use a relatively small modular network. It consists of three layers, containing four components each. The components in the first  layer are 5x5 convolutions, while in the second and third layers are 3x3 convolutions. After the last layer, the output feature map is flattened, and passed through a task-specific linear head.
For more details, see Appendix~\ref{appendix:mnist_mt_network}.

\paragraph{Results}

\begin{table}[t]
\caption{Results on the 4-MNISTs multi-task setup. Each experiment was run $30$ times, we report mean and standard deviation of the error.}
\label{tab:mnist}
\begin{center}
\begin{tabular}{lcc}
\toprule
\multicolumn{1}{c}{\bf Method}  &\multicolumn{1}{c}{\bf Test accuracy (\%)} &\multicolumn{1}{c}{\bf Active connections (\%)} \\
\midrule
No sharing                      & 93.5 $\pm$ 5.1          & 25      \\
Shared bottom                    & 95.7 $\pm$ 0.5          & 100     \\
\midrule
Ours                   & \bf{96.8 $\pm$ 0.2}     & 96      \\
Ours (budget = 0.75)   & \bf{96.7 $\pm$ 0.3}     & 75      \\
\bottomrule
\end{tabular}
\end{center}
\end{table}

We first run two baselines, corresponding to the `no sharing' and `shared bottom' patterns introduced in Section \ref{sec:positive_negative_transfer}. In this case, `no sharing' corresponds to the $i$-th of the four tasks using only the $i$-th component in every layer - again, this means that there is no interaction between tasks. `Shared bottom' means that all tasks use all components. We also train two variants of Gumbel-Matrix: one without any auxiliary penalties, and one with the budget constraint set to $0.75$.
We report all of the results in Table~\ref{tab:mnist}.

We see that `shared bottom' strongly outperforms `no sharing', which shows that this network is small even for MNIST, and using one component per layer is not enough to reliably learn the task.
For the Gumbel-Matrix experiments, we found that the two copies of MNIST end up using the same allocation patterns, as well as the two are copies of MNIST-rot. However, patterns used by the copies of MNIST are different from the ones used by MNIST-rot. As seen in the results, this gives better performance, since the processing is task-dependent. Furthermore, we see that by using the budget penalty, we can reduce the number of active connections without sacrificing the test accuracy.

\subsection{Omniglot}\label{subsec:omniglot}
Next, we test our method on the Omniglot multi-task setup \cite{OmniglotMT-2015}.
The Omniglot dataset consists of $50$ different alphabets, each containing some number of characters. Input samples for each of the characters are handwritten grayscale images of size $105 \times 105$.

\paragraph{Experimental setup}

We create our setup following \cite{MeyerMiikk-ICLR-2018}, where each alphabet is treated as a separate task of predicting the character class. We follow previous works \cite{Liang-2018, Diversity&Depth-ICLR-2019} and use a fixed random subset of $20$ alphabets, splitting every alphabet into training/validation/test sets with proportions $50\%/20\%/30\%$. These alphabets are the first $20$ in the order originally reported by \cite{MeyerMiikk-ICLR-2018}.

In order to have a direct comparison with the state-of-the-art of \cite{Diversity&Depth-ICLR-2019}, we use the same underlying network, optimizer, and regularization techniques; the only difference is how the parameter sharing pattern is learned. We have reached out to the authors of \cite{Diversity&Depth-ICLR-2019} to obtain various details and make sure the setups match. We report the architecture here for completeness.

\begin{figure}[t]
\begin{center}
\includegraphics[width=\linewidth]{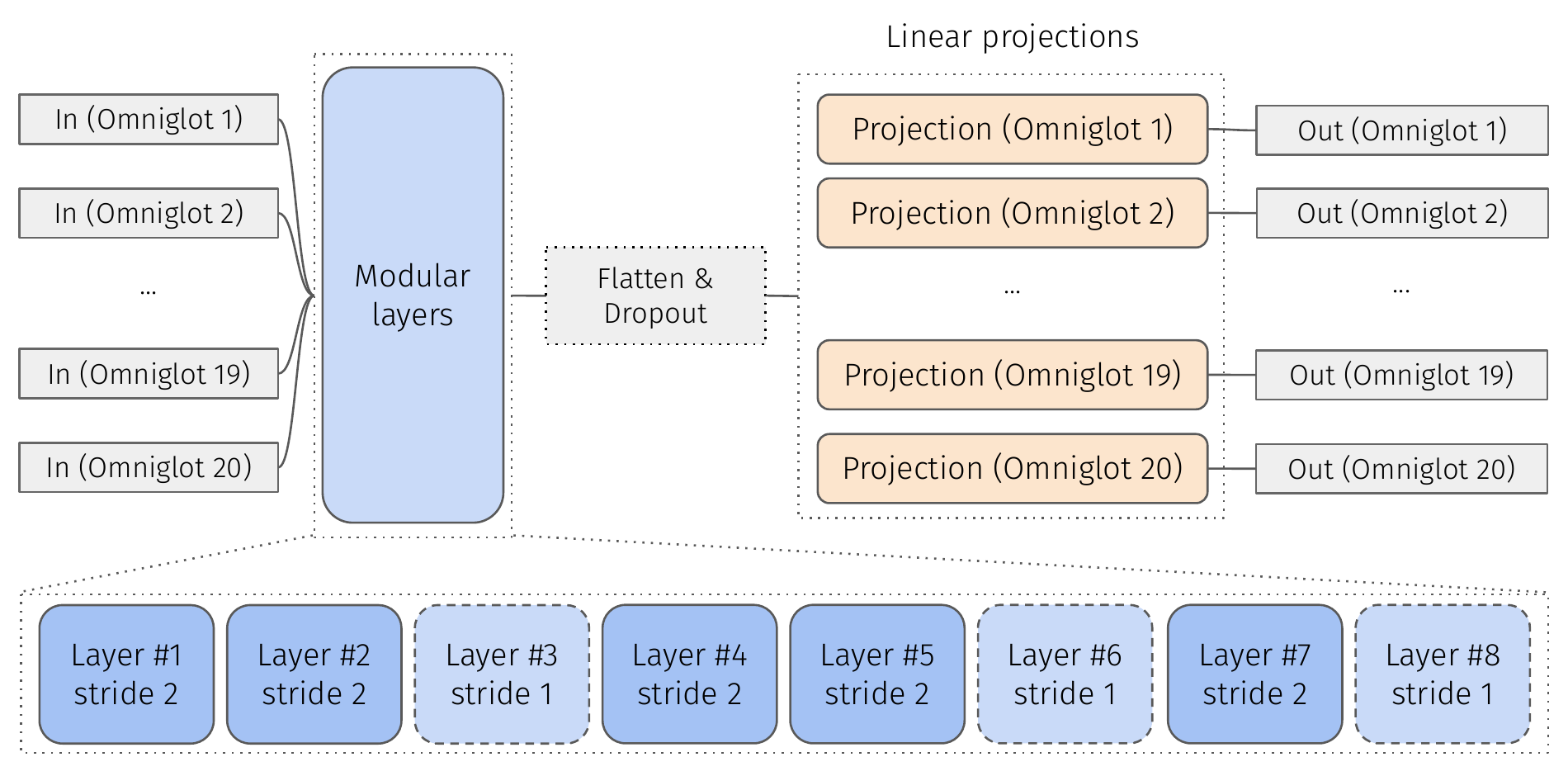}
\end{center}
\caption{The Omniglot multi-task network.}\label{fig:omniglot-model}
\end{figure}

\begin{figure}[t]
\begin{center}
\includegraphics[width=\linewidth]{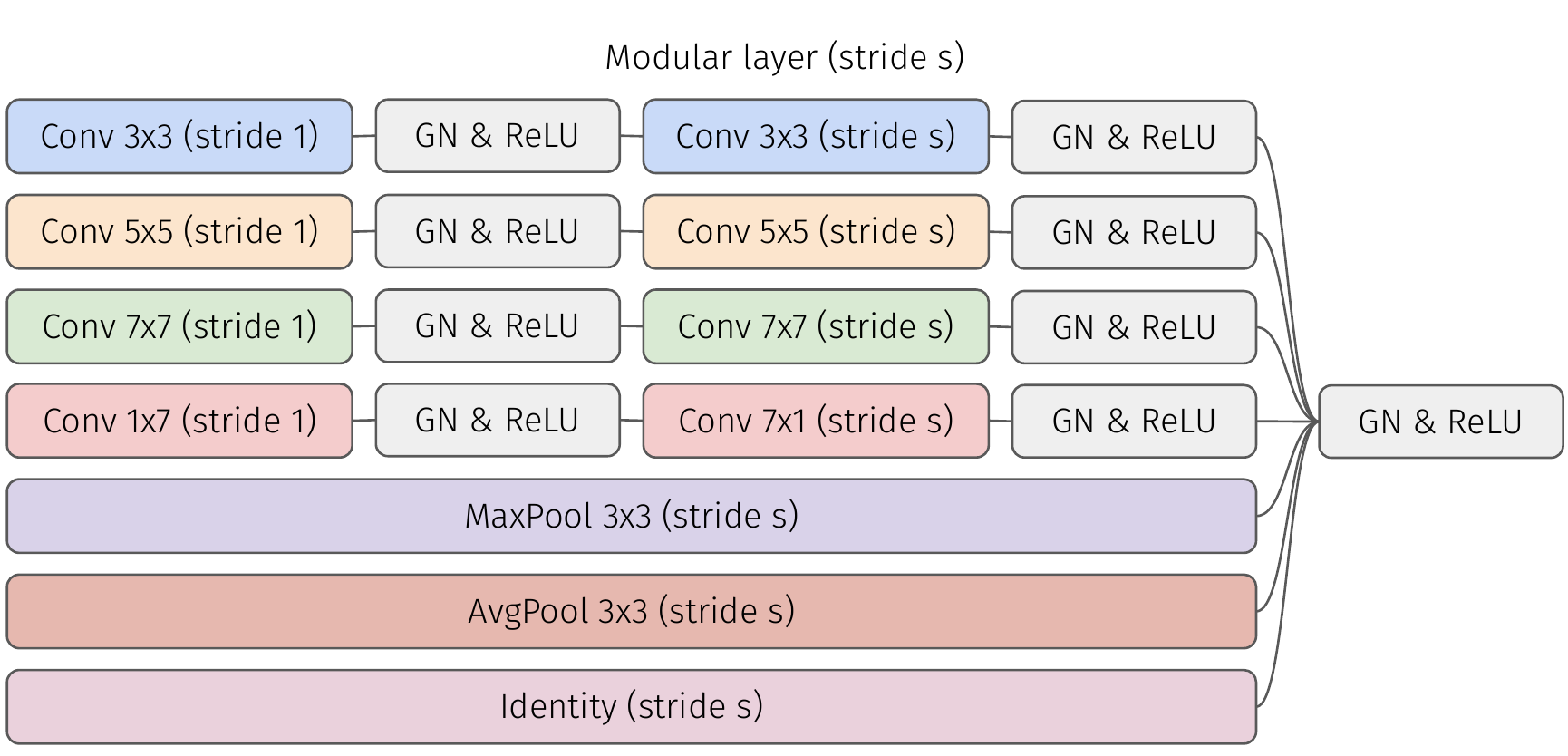}
\end{center}
\caption{Components inside a modular layer in the Omniglot multi-task network. We denote GroupNorm by GN, and the layer stride as $s$. Note that for this specific architecture we have $s \in \{1, 2\}$.}\label{fig:omniglot-layer}
\end{figure}

The network consists of: one shared 1x1 convolution, then $8$ modular layers, and finally linear task specific heads. Each modular layer contains $7$ different components: conv 3x3 $\rightarrow$ conv 3x3, conv 5x5 $\rightarrow$ conv 5x5, conv 7x7 $\rightarrow$ conv 7x7, conv 1x7 $\rightarrow$ conv 7x1, 3x3 max pooling, 3x3 average pooling and identity. The number of channels is $48$ throughout the network. All components use padding to make sure the output shape is the same as the input shape; the spatial dimensions are reduced by adding a stride of $2$ to $5$ of the modular layers. We use GroupNorm \cite{GroupNorm} and ReLU after each convolution and after each modular layer. We present the overview of the architecture in Figure~\ref{fig:omniglot-model}, while a single modular layer is shown in Figure~\ref{fig:omniglot-layer}.

We regularize the model with Dropout and L2-regularization. For training, we use the Adam optimizer. Since the allocation logits are updated only once every $T$ steps (where $T$ is the number of tasks), we have found that for $T = 20$ it is beneficial to use a higher learning rate for updating the allocation logits than that used for updating the components. Therefore, we set the learning rate for the allocation logits to be $T$ times larger than the one for the other model weights, and found this rule of thumb to work reasonably well in practice. We set the training period to be larger than needed for the methods to attain their peak performance, select the best checkpoint for each method based on validation accuracy, and evaluate that single checkpoint on the test set. For hyperparameter values, as well as more details about the network, please see Appendix~\ref{appendix:omniglot_mt_network}.

\paragraph{Results}

Before training a model based on the proposed Gumbel-Matrix method, we train a `shared bottom' variant, where all tasks use all components. We do not evaluate a `no sharing' variant, since the number of tasks $T$ is higher than the number of components per layer.
We show the results of the experiments in Table~\ref{tab:omniglot}.

We see that the `shared bottom' approach actually outperforms the Mixture-of-Experts routing of \cite{Diversity&Depth-ICLR-2019}, which we found to be quite surprising. We conjecture that even though the Mixture-of-Experts variant is more powerful, in the case of multi-task learning on Omniglot, optimization difficulties outweigh that benefit. In contrast to our method, the Mixture-of-Experts framework hard-codes the required sparsity for each layer. This can bring immense computational savings \cite{Shazeer-MoE-2017}, but may also sacrifice accuracy. In some cases, like the one of \cite{Shazeer-MoE-2017}, the `shared bottom' variant would be prohibitively expensive to run, making the comparison infeasible. However, we encourage researchers to compare their methods with their `shared bottom' counterparts whenever possible.

Next, we train the model based on the proposed Gumbel-Matrix method. We do not use any auxiliary losses, and find that the model naturally removes some of the allocations to allow for task-specific processing. Even though the network is not explicitly penalized for high entropy, the allocation probabilities still converge to be either close to $0.0$ or close to $1.0$. We report the resulting accuracy in Table~\ref{tab:omniglot}. We see that the Gumbel-Matrix method improves the accuracy over a very strong `shared bottom' baseline.

\paragraph{Analysis}

In order to analyze the parameter sharing patterns learned by our network, we compute several statistics. First, for each task, we compute the binary embedding vector according to Section~\ref{subsec:embeddings}. Note that if two tasks have exactly the same embedding, the network will process them in the same way (up to the task-specific linear heads). We find that, on average, the model converges to having $10.1$ different embedding vectors. That means that tasks are `clustered' into groups that are processed in the same way, and the average size of such group is approximately $2$.

Next, we analyze what kinds of components are used at every depth in the network. Recall that each layer in the model consists of $7$ components, out of which $4$ are different kinds of convolutions, $2$ are pooling (max and average), and the last one is identity, which can be thought of as a skip connection. We found that the usage trends are consistent across experimental runs, and mostly depend on the layer in question. The trends are the following (please see Appendix~\ref{sec:prob_over_time_vis} for detailed illustrations):
\begin{itemize}
    \item In layers $1-4$, we found that the allocations of all components are being dropped at a similar rate; most of the diversity in processing different tasks concentrates in these layers.
    \item In layers $5-7$, almost all components are always used, and the only source of differences between tasks is that some of them use the average pooling component, while others don't.
    \item In layer $8$, the average pooling component is never used, and the other components are always used.
\end{itemize}

Our interpretation of this result is that the Gumbel-Matrix shows a mixture of two behaviors. First, components can be dropped in order to obtain task-specific processing; this trend is visible in the first layers of the model. Second, all allocations to specific components can be removed, which corresponds to a form of network pruning. We speculate that the task-specific processing in the initial layers `aligns' the feature spaces of samples coming from different tasks. In the final layers, the tasks are processed in a uniform way, but some components are pruned to improve the model quality.

\begin{table}[tb]
\caption{Results on multi-task Omniglot setup. Each experiment was run $10$ times, we report mean and standard deviation of the error.}
\label{tab:omniglot}
\begin{center}
\begin{tabular}{lcc}
\toprule
\multicolumn{1}{c}{\bf Method}  &\multicolumn{1}{c}{\bf Valid. error (\%)} &\multicolumn{1}{c}{\bf Test error (\%)} \\
\midrule
Single Task {\cite{MeyerMiikk-ICLR-2018}}   & 36.41 $\pm$ 0.53          & 39.19 $\pm$ 0.50      \\
Soft Ordering {\cite{MeyerMiikk-ICLR-2018}} & 32.33 $\pm$ 0.74          & 33.41 $\pm$ 0.71      \\
CMTR {\cite{Liang-2018}}                    & 11.80 $\pm$ 1.02          & 12.81 $\pm$ 1.02      \\
MoE {\cite{Diversity&Depth-ICLR-2019}}      & 7.95 $\pm$ 0.37           & 7.81 $\pm$ 0.54       \\
\midrule
Shared bottom                                 & 6.16 $\pm$ 0.50           & 6.75 $\pm$ 0.33       \\
Gumbel-Matrix                                & \bf{5.69 $\pm$ 0.22}      & \bf{6.48 $\pm$ 0.28}  \\
\bottomrule
\end{tabular}
\end{center}
\end{table}

\subsection{Discovering latent clusters of related tasks}\label{subsec:clusters}

To test our hypothesis from Section~\ref{subsec:embeddings} that the proposed method can discover the task relatedness, we create a setup in which there are clearly defined latent clusters of tasks.
Specifically, we use a benchmark composed of three sets of tasks:
\begin{itemize}
    \item Following \cite{RoutingNetworks18}, we create $20$ tasks from the CIFAR-100 dataset \cite{krizhevsky2009learning}. For each of the $20$ coarse labels, we create a task of predicting the fine-grained label.
    \item We create $10$ variants of the MNIST dataset; the $i$-th variant is the task of telling apart all MNIST digits besides digit $i$. In other words, every MNIST variant is a $9$-way classification task, and the $i$-th variant never sees nor is asked to classify the digit $i$.
    \item In the same way as with MNIST, we create $10$ variants of the Fashion-MNIST dataset \cite{xiao2017fashion}.
\end{itemize}

This benchmark contains $T = 40$ tasks. Note that the tasks within the MNIST cluster are highly related, since any two tasks within the cluster share $8$ out of $9$ of the classification classes. Similarly, tasks within the Fashion-MNIST cluster are highly related. On the other hand, tasks within the CIFAR-100 cluster are related since the input images come from the same data distribution, but the shared information is more on the level of features than on the level of classes. 
Note that the Gumbel-Matrix framework is informed with only the task ids ranging from $0$ and $T-1$, and has no access to information regarding the task set composition.

\begin{figure}[t]
\begin{center}
\includegraphics[width=0.8\linewidth]{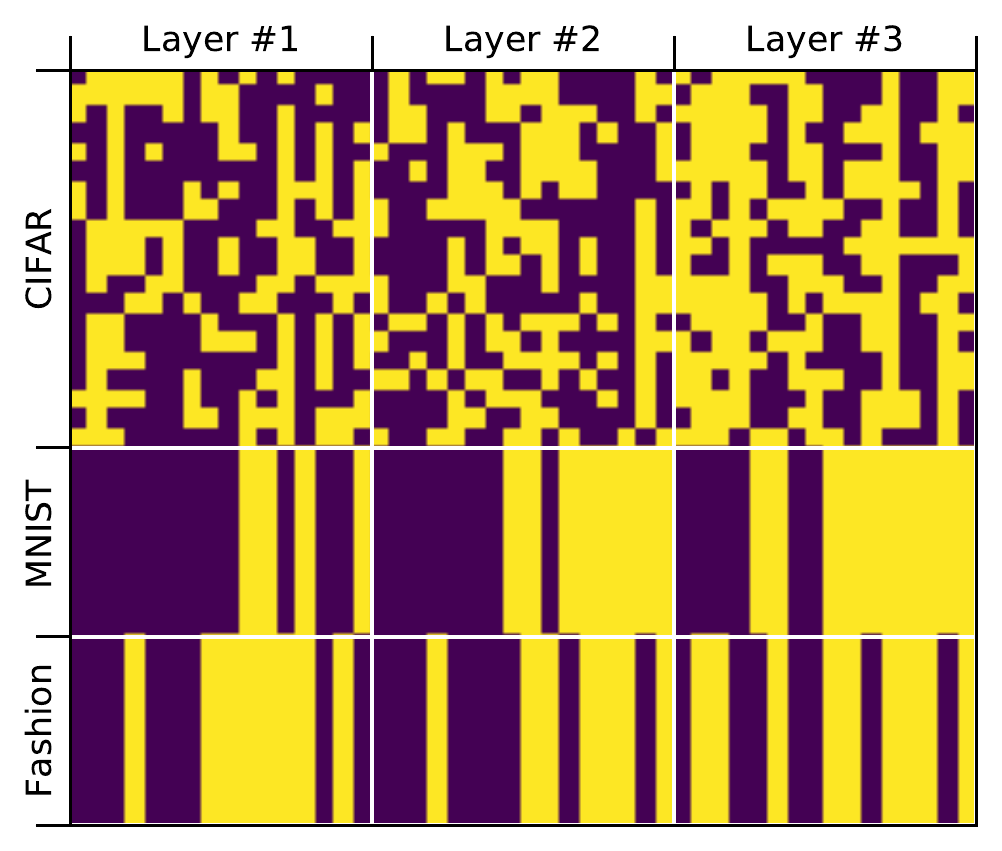}
\end{center}
\caption{Binary allocation vectors for the $40$ tasks. Rows correspond to tasks: first $20$ rows form the CIFAR cluster, next $10$ the MNIST cluster, and the last $10$ the Fashion-MNIST cluster. Columns correspond to the $48$ components of the model: $16$ components in each of the $3$ modular layers. A yellow pixel denotes a $1$ (the component is allocated to a given task), while a purple pixel denotes a $0$.}\label{fig:pattern}
\end{figure}

\paragraph{Experimental setup}
The input images for each of the MNIST and Fashion-MNIST tasks are in grayscale, while CIFAR contains RGB images.
In order for the input shape to match across all tasks, we reshape the images from the MNIST and Fashion-MNIST  from $28$x$28$ to $32$x$32$. We also convert these images to have three channels instead of one.

The architecture of the neural network used in the following experiments consists of: one shared 1x1 convolution, then $3$ modular layers, and finally linear task specific heads. Each modular layer contains $16$ different components: $8$ 3x3 convolutions, and $8$ 5x5 convolutions. The number of channels throughout the network is $8$. We follow the setup described in Section~\ref{subsec:omniglot} with respect to the use of padding, GroupNorm and ReLU. All modular layers downscale the input by applying a stride of $2$.

Finally, for all tasks, we employ a standard CIFAR image augmentation: we pad the $32$x$32$ input image to the size of $40$x$40$, and then crop out a random region of shape $32$x$32$.

We train with Adam optimizer, with learning rate set to $0.0003$ for the model parameters, and to $0.006$ for the allocation logits. We also use a budget of $b = 0.5$ with $\lambda = 1$ to promote sparsity. The allocation probabilities are initialized to $p_{init} = 0.5$, matching the budget.

\paragraph{Analysis}

After training with the Gumbel-Matrix method, we extract binary allocation vectors of length $C = 3 \cdot 16 = 48$ for each of the tasks, following Section~\ref{subsec:embeddings}.
We visualize the embedding vectors in Figure~\ref{fig:pattern}. Interestingly, all tasks within the MNIST cluster are assigned the same embedding, matching the expectation that all the tasks in this cluster are highly related. The same holds for tasks within the Fashion-MNIST cluster. On the other hand, tasks within the CIFAR cluster are assigned varying embeddings, hinting that this cluster is more diverse.

\begin{figure}[t]
\begin{center}
\includegraphics[width=0.68\linewidth]{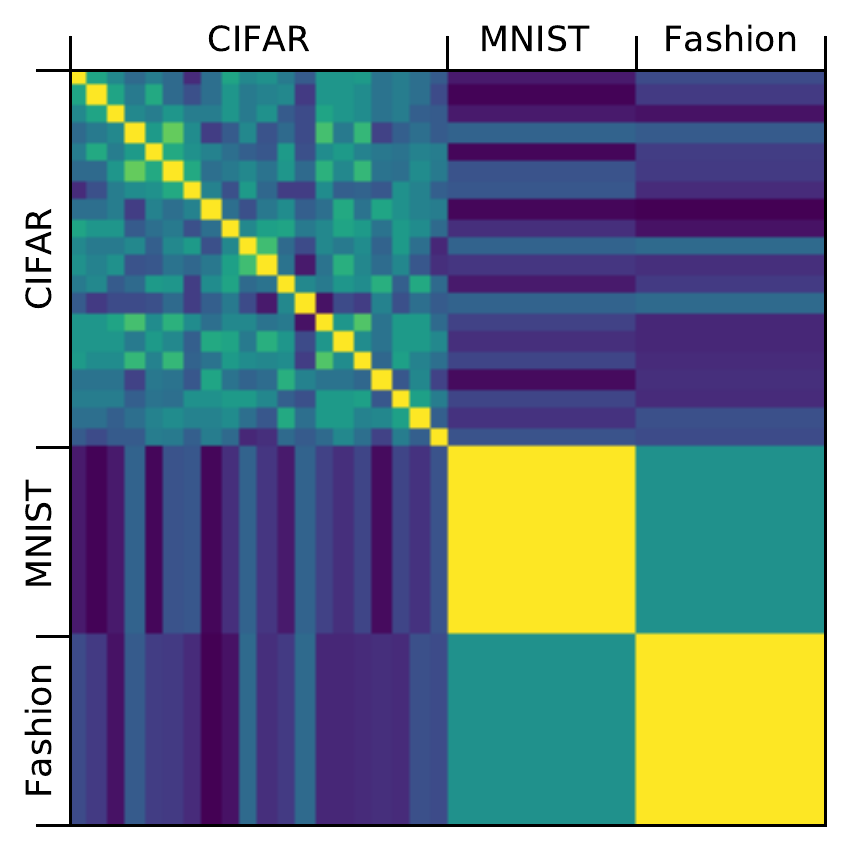}
\end{center}
\caption{Pairwise cosine similarity of the binary embedding vectors for the $40$ tasks. Rows and columns correspond to tasks. Yellow pixels denote similarity of $1$ (vectors are the same), while on the other end of the spectrum purple denote similarity of $0$ (vectors are orthogonal).}\label{fig:similarity}
\end{figure}

To further illustrate that the method can recover the structure of the three clusters, we compute the pairwise cosine similarity for each pair of task embeddings, and visualize the resulting distance matrix in Figure~\ref{fig:similarity}. As noted before, the CIFAR cluster seems to be the most inhomogeneous, but remains distinct from the other two clusters. Moreover, the MNIST and Fashion-MNIST clusters are closer to each other than to the CIFAR cluster, which may be related to the fact that these two clusters have grayscale input images.

\section{Conclusions and future work}\label{sec:conclusions}
We introduced a new method for multi-task learning that learns the pattern of parameter sharing together with the model parameters using standard back-propagation. We provided experimental results showing that our method can learn flexible sharing patterns that leverage task relatedness, which results in significantly improved performances over the state-of-the-art.
The presented method conditions the binary allocation variables only on the task id. Future work can extend this method by conditioning the allocation variables on richer information such as task metadata.

\bibliography{references}

\begin{thebibliography}{10}
\providecommand{\url}[1]{#1}
\csname url@samestyle\endcsname
\providecommand{\newblock}{\relax}
\providecommand{\bibinfo}[2]{#2}
\providecommand{\BIBentrySTDinterwordspacing}{\spaceskip=0pt\relax}
\providecommand{\BIBentryALTinterwordstretchfactor}{4}
\providecommand{\BIBentryALTinterwordspacing}{\spaceskip=\fontdimen2\font plus
\BIBentryALTinterwordstretchfactor\fontdimen3\font minus
  \fontdimen4\font\relax}
\providecommand{\BIBforeignlanguage}[2]{{%
\expandafter\ifx\csname l@#1\endcsname\relax
\typeout{** WARNING: IEEEtran.bst: No hyphenation pattern has been}%
\typeout{** loaded for the language `#1'. Using the pattern for}%
\typeout{** the default language instead.}%
\else
\language=\csname l@#1\endcsname
\fi
#2}}
\providecommand{\BIBdecl}{\relax}
\BIBdecl

\bibitem{Caruana-1998}
R.~Caruana, ``{Multitask learning},'' \emph{Learning to learn}, 1998.

\bibitem{Caruana-1993}
------, ``{Multitask learning: A knowledge-based source of inductive bias},''
  \emph{In Machine Learning: Proceedings of the Tenth International
  Conference}, pp. 41--48, 1993.

\bibitem{BaBeCa-2016}
T.~Bansal, D.~Belanger, and A.~McCallum, ``{Ask the GRU: Multi-task learning
  for deep text recommendations},'' in \emph{10th ACM Conference on Recommender
  Systems}.\hskip 1em plus 0.5em minus 0.4em\relax ACM, 2016, pp. 107–--114.

\bibitem{Fast-R-CNN}
R.~Girshick, ``{Fast R-CNN},'' in \emph{International Conference on Computer
  Vision (ICCV)}.\hskip 1em plus 0.5em minus 0.4em\relax IEEE, 2015, pp.
  1440–--1448.

\bibitem{DBLP:journals/corr/abs-1907-04829}
\BIBentryALTinterwordspacing
K.~Clark, M.~Luong, U.~Khandelwal, C.~D. Manning, and Q.~V. Le, ``Bam!
  born-again multi-task networks for natural language understanding,''
  \emph{CoRR}, vol. abs/1907.04829, 2019. [Online]. Available:
  \url{http://arxiv.org/abs/1907.04829}
\BIBentrySTDinterwordspacing

\bibitem{DBLP:journals/corr/abs-1907-12412}
\BIBentryALTinterwordspacing
Y.~Sun, S.~Wang, Y.~Li, S.~Feng, H.~Tian, H.~Wu, and H.~Wang, ``{ERNIE} 2.0:
  {A} continual pre-training framework for language understanding,''
  \emph{CoRR}, vol. abs/1907.12412, 2019. [Online]. Available:
  \url{http://arxiv.org/abs/1907.12412}
\BIBentrySTDinterwordspacing

\bibitem{JangGuPoole.2017}
E.~Jang, S.~Gu, and B.~Poole, ``{Categorical reparametrization with
  Gumbel-softmax},'' \emph{International Conference on Learning Representations
  (ICLR)}, 2017.

\bibitem{MMoE-KDD-2018}
J.~Ma, Z.~Zhao, Y.~X., J.~Chen, L.~Hong, and E.~Chi, ``{Modeling Task
  Relationships in Multi-task Learning with Multi-gate Mixture-of-Experts},''
  \emph{KDD}, 2018.

\bibitem{SpotTune-CVPR-2018}
\BIBentryALTinterwordspacing
Y.~Guo, H.~Shi, A.~Kumar, K.~Grauman, T.~Rosing, and R.~Feris, ``{SpotTune:
  Transfer Learning through Adaptive Fine-tuning},'' \emph{Conference on
  Computer Vision and Pattern Recognition (CVPR)}, 2018. [Online]. Available:
  \url{https://arxiv.org/abs/1811.08737}
\BIBentrySTDinterwordspacing

\bibitem{RoutingNetworks18}
\BIBentryALTinterwordspacing
C.~Rosenbaum, T.~Klinger, and M.~Riemer, ``{Routing Networks: Adaptive
  Selection of Non-Linear Functions for Multi-Task Learning},''
  \emph{International Conference on Learning Representations (ICLR)}, 2018.
  [Online]. Available: \url{https://openreview.net/pdf?id=ry8dvM-R-}
\BIBentrySTDinterwordspacing

\bibitem{RoutingNetsChallenges}
C.~Rosenbaum, I.~Cases, M.~Riemer, and T.~Klinger, ``Routing networks and the
  challenges of modular and compositional computation,'' \emph{arXiv preprint
  arXiv:1904.12774}, 2019.

\bibitem{Shazeer-MoE-2017}
N.~Shazeer, A.~Mirhoseini, K.~Maziarz, A.~Davis, Q.~Le, G.~Hinton, and J.~Dean,
  ``Outrageously large neural networks: The sparsely-gated mixture-of-experts
  layer,'' \emph{arXiv preprint arXiv:1701.06538}, 2017.

\bibitem{Diversity&Depth-ICLR-2019}
\BIBentryALTinterwordspacing
P.~Ramachandran and Q.~V. Le, ``{Diversity and Depth in Per-Example Routing
  Models},'' \emph{International Conference on Learning Representations
  (ICLR)}, 2019. [Online]. Available:
  \url{https://openreview.net/pdf?id=BkxWJnC9tX}
\BIBentrySTDinterwordspacing

\bibitem{Fernando2017-Pathnet}
C.~Fernando, D.~Banarse, C.~Blundell, Y.~Zwols, D.~Ha, A.~A. Rusu, A.~Pritzel,
  and D.~Wierstra, ``Pathnet: Evolution channels gradient descent in super
  neural networks,'' \emph{arXiv preprint arXiv:1701.08734}, 2017.

\bibitem{CrossStichNetworks}
I.~Misra, A.~Shrivastava, A.~Gupta, , and M.~Hebert, ``{Cross-stitch networks
  for multi-task learning},'' in \emph{Conference on Computer Vision and
  Pattern Recognition}.\hskip 1em plus 0.5em minus 0.4em\relax IEEE, 2016, pp.
  3994–--4003.

\bibitem{MeyerMiikk-ICLR-2018}
E.~Meyerson and R.~Miikkulainen, ``{Beyond shared hierarchies: Deep multitask
  learning through soft layer ordering},'' \emph{International Conference on
  Learning Representations (ICLR)}, 2018.

\bibitem{SNR-AAAI19}
\BIBentryALTinterwordspacing
J.~Ma, Z.~Zhao, J.~Chen, A.~Li, L.~Hong, and E.~Chi, ``{SNR: Sub-Network
  Routing for Flexible Parameter Sharing in Multi-task Learning},'' \emph{AAAI
  Conference on Artificial Intelligence}, 2019. [Online]. Available:
  \url{http://www.jiaqima.com/papers/SNR.pdf}
\BIBentrySTDinterwordspacing

\bibitem{ManyTaskLearning}
G.~Strezoski, N.~van Noord, and M.~Worring, ``{Many Task Learning With Task
  Routing},'' in \emph{Proceedings of the IEEE International Conference on
  Computer Vision}, 2019, pp. 1375--1384.

\bibitem{BengBaPiPre.2016}
E.~Bengio, P.-L. Bacon, J.~Pineau, and D.~Precup, ``Conditional computation in
  neural networks for faster models,'' \emph{arXiv preprint arXiv:1511.06297},
  2016.

\bibitem{REINFORCE-1992}
R.~J. Williams, ``Simple statistical gradient-following algorithms for
  connectionist reinforcement learning,'' in \emph{Machine Learning}, vol.
  8(3--4), 1992, pp. 229–--256.

\bibitem{ConvNet-AIG-19}
A.~Veit and S.~Belongie, ``{Convolutional Networks with Adaptive Inference
  Graphs},'' in \emph{International Journal of Computer Vision (IJCV)}, 2019,
  pp. 1–--12.

\bibitem{ZophLe16}
B.~Zoph and Q.~V. Le, ``{Neural Architecture Search with Reinforcement
  Learning},'' \emph{International Conference on Learning Representations
  (ICLR)}, 2017.

\bibitem{ReMoSeSaSuTaLeKu17}
E.~Real, S.~Moore, A.~Selle, S.~Saxena, Y.~L. Suematsu, J.~Tan, Q.~V. Le, and
  A.~Kurakin, ``{Large-Scale Evolution of Image Classifiers},''
  \emph{International Conference on Machine Learning (ICML)}, 2017.

\bibitem{maziarz2018evolutionary}
K.~Maziarz, M.~Tan, A.~Khorlin, M.~Georgiev, and A.~Gesmundo,
  ``Evolutionary-neural hybrid agents for architecture search,''
  \emph{International Conference on Machine Learning (ICML): Workshop on
  AutoML}, 2018.

\bibitem{white2019bananas}
C.~White, W.~Neiswanger, and Y.~Savani, ``Bananas: Bayesian optimization with
  neural architectures for neural architecture search,'' \emph{arXiv preprint
  arXiv:1910.11858}, 2019.

\bibitem{ENAS}
H.~Pham, M.~Guan, B.~Zoph, Q.~Le, and J.~Dean, ``Efficient neural architecture
  search via parameters sharing,'' \emph{International Conference on Machine
  Learning (ICML)}, 2018.

\bibitem{DARTS}
H.~Liu, K.~Simonyan, and Y.~Yang, ``{DARTS}: Differentiable architecture
  search,'' in \emph{International Conference on Learning Representations
  (ICLR)}, 2019.

\bibitem{OmniglotMT-2015}
M.~B. Lake, R.~Salakhutdinov, and J.~B. Tenenbaum, ``{Human-level concept
  learning through probabilistic program induction},'' \emph{Science}, vol.
  350, p. 1332–1338, 2015.

\bibitem{Liang-2018}
J.~Liang, E.~Meyerson, and R.~Miikkulainen, ``Evolutionary architecture search
  for deep multitask networks,'' in \emph{Proceedings of the Genetic and
  Evolutionary Computation Conference}.\hskip 1em plus 0.5em minus 0.4em\relax
  ACM, 2018, pp. 466--473.

\bibitem{GroupNorm}
Y.~Wu and K.~He, ``Group normalization,'' in \emph{Proceedings of the European
  Conference on Computer Vision (ECCV)}, 2018, pp. 3--19.

\bibitem{krizhevsky2009learning}
A.~Krizhevsky \emph{et~al.}, ``Learning multiple layers of features from tiny
  images,'' 2009.

\bibitem{xiao2017fashion}
H.~Xiao, K.~Rasul, and R.~Vollgraf, ``Fashion-mnist: a novel image dataset for
  benchmarking machine learning algorithms,'' \emph{arXiv preprint
  arXiv:1708.07747}, 2017.

\end{thebibliography}
\bibliographystyle{IEEEtran}

\appendix

\begin{figure*}[t]
\begin{center}
\includegraphics[width=0.75\linewidth]{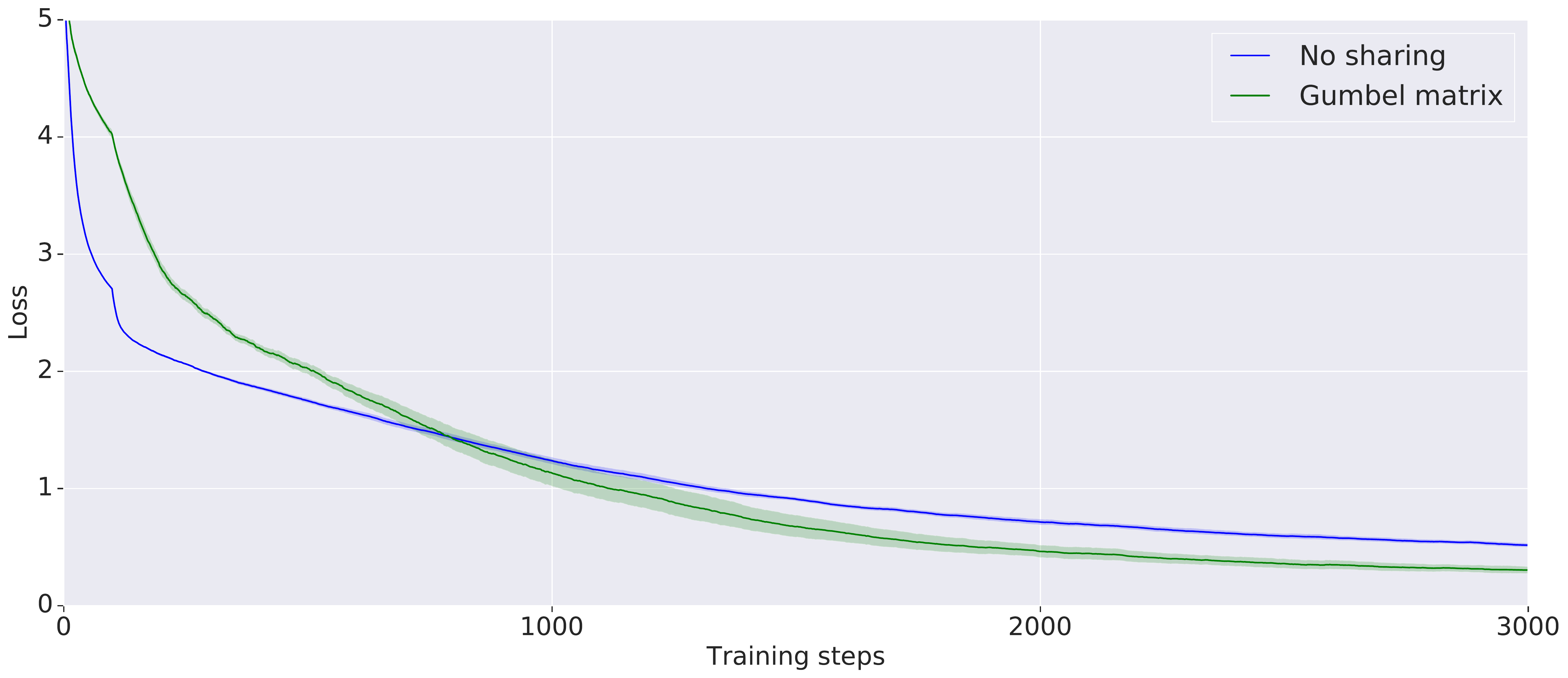}
\end{center}
\caption{Comparison of the `no sharing' pattern with the Gumbel-Matrix method on the synthetic data experiment. The plot shows loss over time (averaged over the four tasks and smoothed over a window of $100$ steps). We ran each experiment $20$ times, and the shaded area corresponds to the $90\%$ confidence interval.}\label{fig:gumbel_matrix_synthetic_data}
\end{figure*}

\subsection{Synthetic data}

\subsubsection{Generation}\label{appendix:synthetic_data_generation}

We follow a process for synthetic data generation based on \cite{MMoE-KDD-2018}, which allows for explicit control of the relatedness between tasks. In particular, we generate the data as follows:
\begin{enumerate}
\item We generate two orthogonal unit vectors $u_1, u_2 \in R^d$. That is, $u_1$ and $u_2$ satisfy $u_1 ^\top u_2 = 0$ and $\|u_1\| = \|u_2\| = 1$.
\item Given a desired score of task relatedness $0 \leq \rho \leq 1 $, we generate two weight vectors $w_1, w_2$ such that
\begin{equation}
    w_1 = c u_1, w_2 = c(\rho u_1 + \sqrt{1-\rho^2} u_2)
\end{equation}
\item We randomly sample a data point $x \in R^d$ where each dimension is sampled from $\mathcal{N}(0,1)$.
\item Generate two labels $y_1, y_2$ for the two tasks as follows:
\begin{equation}
y_1 = w_1^\top x + \sum_{i=1}^m \sin(\alpha_i w_1^\top x + \beta_i) + \epsilon_1
\end{equation}
\begin{equation}
y_2 = w_2^\top x + \sum_{i=1}^m \sin(\alpha_i w_2^\top x + \beta_i) + \epsilon_2
\end{equation}
where $\epsilon_1, \epsilon_2 \sim \mathcal{N}(0, 0.01)$.
\end{enumerate}

In the procedure above, the values of $d$, $c$, $m$, and the sequences $\alpha_i$, $\beta_i$ are hyperparameters. For all of our experiments with synthetic data, we set $d = 128$, $c=1$, $m = 6$, $\alpha_i = i$, $\beta_i = (i - 1)^2$. We found $m$ to be the most important of these hyperparameters, as it controls the non-linearity (and hence the difficulty) of the tasks. In particular, for $m = 0$ the input-output relation for the produced task will be linear (up to noise).

\subsubsection{Architecture for the `Positive and Negative transfer' experiment}\label{appendix:positive_negative_transfer_arch}

Here we describe in detail the network used for the experiment in Section~\ref{sec:positive_negative_transfer}.

The network starts with a layer containing $4$ parallel fully connected components. Two of these components components consist of 1 fully connected layer, and the other two consist of 2 layers with $4$ hidden units. All components have $4$ output units, and use ReLU activations on the hidden and output units. Note that a specific input can be passed through multiple components; the outputs of the components are averaged before being passed to a task-specific linear head.

\subsubsection{Gumbel-Matrix experiment}\label{appendix:synthetic_data_gumbel_matrix}

We additionally test the Gumbel-Matrix method on synthetic data. To that end, we create a modular network that does not contain task specific heads. Instead, it consists of a layer with $4$ components, each with $1$ output unit. Each component is $3$ layers deep, with $16$ hidden units in every hidden layer, ReLU activations on the hidden units, and no activation on the output units. We use batch size of $64$, learning rate $0.01$, and we clip the gradient norm to $1.0$.

We generate two pairs of tasks according to Appendix~\ref{appendix:synthetic_data_generation}, so that each pair contains tasks that are the same up to noise, but tasks in different pairs are unrelated. Since we use a network with no task-specific heads, if two tasks go through the same set of components, the network will implement the same function for these tasks. Therefore, if the allocation pattern is incorrect (two tasks from different pairs go through the same subset of components), the rest of the network cannot make up for it, making it a good sanity check for our method.

Note that in this network, each component is capable of learning the synthetic task on its own, and using a `no sharing' pattern is a solid baseline. However, dynamically learning the parameter sharing pattern should still be able to improve performance: if it manages to correctly discover the related pairs of tasks, they can use a shared set of components, and parameters in these components would get twice as much training data compared to the `no sharing' case.

We train for $3000$ batches per task, and track the average loss over tasks. The result is shown in Figure~\ref{fig:gumbel_matrix_synthetic_data}. We found that the related tasks converge to using the same subset of components, but the subsets used by the unrelated tasks are distinct. Because of that, the Gumbel-Matrix method outperforms the `no sharing' variant. In this setup, the `shared bottom' setup works very poorly, since it is unable to learn both pairs of tasks at the same time. Hence, we omit the results for the `shared bottom' baseline.

\subsection{Architecture for the multi-task MNIST experiment}\label{appendix:mnist_mt_network}

Here we report additional details on the network used for the MNIST experiments. 

As stated before, the network consists of three modular layers, containing convolutions with kernels 5x5, 3x3 and 3x3, respectively. We do not use padding in any of the convolutions, which means spatial dimensions are slightly reduced in each modular layer. After the first and second layer, we further reduce the spatial dimensions by applying 2x2 average pooling. All convolutions throughout the network have $4$ filters.

During training, we regularize the model with Dropout (dropout probability $0.5$, applied right before the tasks-specific heads). We set the training length to $10$ epochs per task, and use a batch size of $16$.

\subsection{Omniglot}

\subsubsection{Architecture and hyperparameters}\label{appendix:omniglot_mt_network}

For the Omniglot experiments, we use a large convolutional network, developed by the authors of “Diversity and Depth in Per-Example Routing Models” \cite{Diversity&Depth-ICLR-2019}.

\begin{figure*}[t]
\begin{center}
\includegraphics[width=0.9\linewidth]{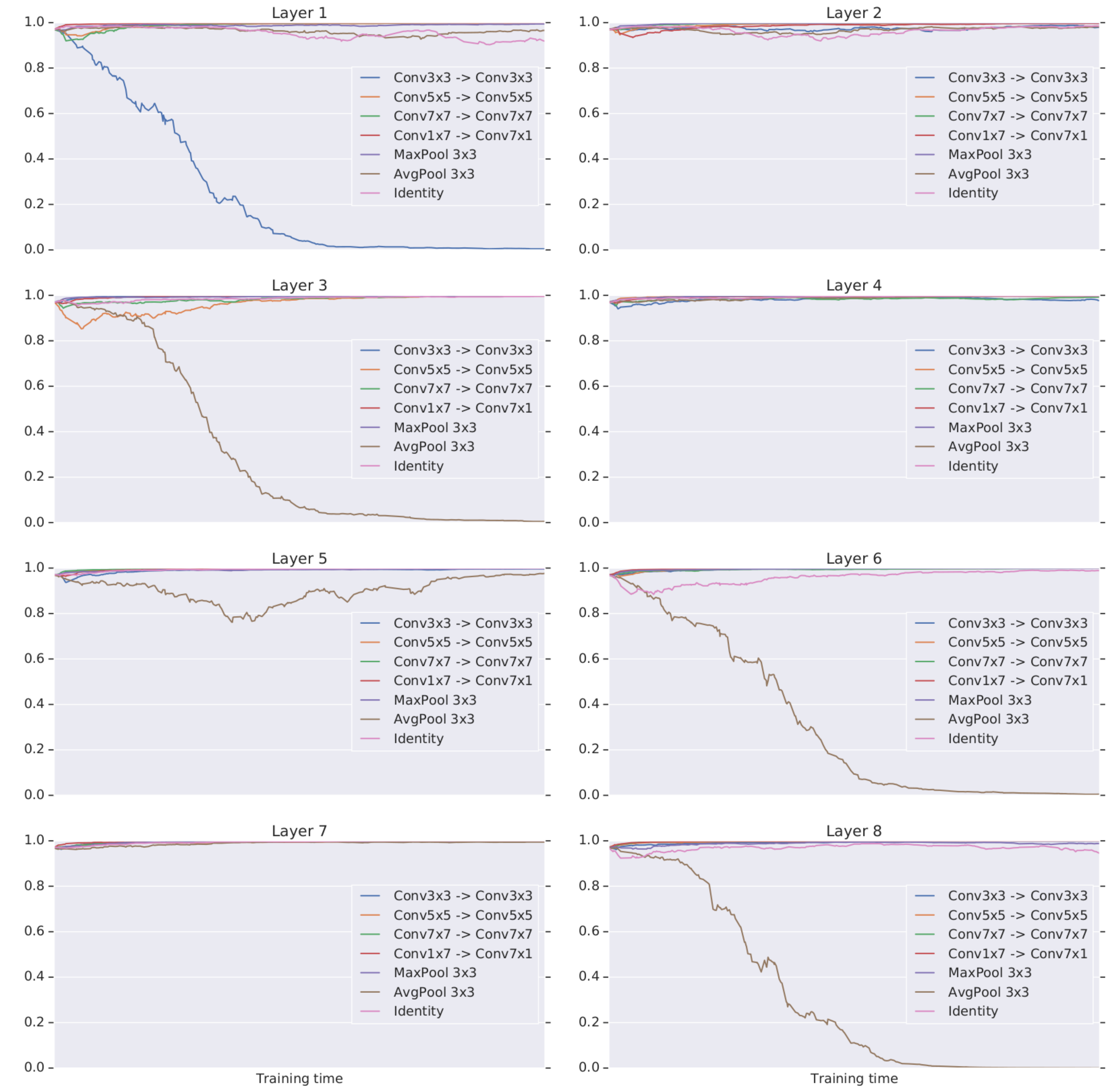}
\end{center}
\caption{Allocation probabilities over time (for one run, and one task). Each plot above shows the allocation probabilities for all components of a certain modular layer in the network.}\label{fig:omniglot_probs_over_time}
\end{figure*}

One detail that we copy from \cite{Diversity&Depth-ICLR-2019} is that the identity component, in the case when it belongs to a  layer with a stride larger than $1$, is actually implemented as a strided 1x1 convolution. While it makes the name `identity component' slightly inappropriate, we follow this to be directly comparable.

We tuned the network hyperparameters using the validation set, and used the following values for all of our Omniglot experiments. Dropout probability is $0.5$, L2-regularization strength $0.0003$, and the learning rate is set to $0.0001$. We train with a batch size of $16$.

For the Gumbel-Matrix method, we set $p_{init} = 0.97$. We found that increasing $p_{init}$ does slightly increase the average number of active connections at convergence, however, even with initialization as high as $0.97$, the model still confidently removes many connections. On the other hand, we found $p_{init} = 0.97$ trains much better than $p_{init} = 0.5$. We believe that in a deep network with many components in every layer, setting $p_{init}$ to $0.5$ introduces too much noise during training.

\subsubsection{Additional analysis}\label{sec:prob_over_time_vis}

As we have discussed in Section~\ref{subsec:omniglot}, the connection to the average pooling component is being dropped most often, while other connections are only dropped in the early layers.

Figure~\ref{fig:omniglot_probs_over_time} illustrates these trends and shows how the connection probabilities are evolving over time for a the `Syriac' alphabet task. Note that for the experiments in Section~\ref{subsec:omniglot} we ran $10$ experiments for every method; here, for ease of presentation, we only show one of the runs. We see for example that for this specific task, the connection to the $3 \times 3$ convolutional component gets dropped in the first layer, while connections to average pooling are dropped in layers 3, 6 and 8. In the remaining three layers, the task uses all components.

\end{document}